\documentclass[10pt,journal,cspaper,compsoc]{IEEEtran}

\ifCLASSOPTIONcompsoc
\else
\fi

\usepackage{graphicx}
\graphicspath{{./Figs/}}
\DeclareGraphicsExtensions{.pdf}

\usepackage{amssymb}
\usepackage[cmex10]{amsmath}

\usepackage{url}
\usepackage{epstopdf}
\usepackage{pdfpages}
\usepackage{colortbl}
\usepackage[ruled,vlined]{algorithm2e}
\usepackage{subfigure}
\begin{document}


\title{A Hash-based Co-Clustering Algorithm for Categorical Data}

\author{Fabr\'icio~Olivetti~de~Fran\c{c}a,~\IEEEmembership{Member,~IEEE,}
\IEEEcompsocitemizethanks{\IEEEcompsocthanksitem F. O. de Fran\c{c}a is with the Center of Mathematics, Computing and Cognition (CMCC), Universidade Federal do ABC (UFABC) -- Santo Andr\'{e}, SP, Brazil.\protect\\
E-mail: folivetti@ufabc.edu.br}
\thanks{}}

\IEEEcompsoctitleabstractindextext{%
\begin{abstract}
  Many real-life data are described by categorical attributes without a pre-classification. A common data mining method used to extract information from this type of data is clustering. This method group together the samples from the data that are more similar than all other samples. But, categorical data pose a challenge when extracting information because: the calculation of two objects' similarity is usually done by measuring the number of common features, but ignore a possible importance weighting; if the data may be divided differently according to different subsets of the features, the algorithm may find clusters with different meanings from each other, difficulting the post analysis. Data Co-Clustering of categorical data is the technique that tries to find subsets of samples that share a subset of features in common. By doing so, not only a sample may belong to more than one cluster but, the feature selection of each cluster describe its own characteristics. In this paper a novel Co-Clustering technique for categorical data is proposed by using Locality Sensitive Hashing technique in order to preprocess a list of Co-Clusters seeds based on a previous research. Results indicate this technique is capable of finding high quality Co-Clusters in many different categorical data sets and scales linearly with the data set size.
\end{abstract}

\begin{keywords}

co-clustering, categorical data, data mining, text mining, biclustering
\end{keywords}
}
\maketitle
\IEEEdisplaynotcompsoctitleabstractindextext
\IEEEpeerreviewmaketitle

% \linenumbers

\section{Introduction}
\label{sec:intro}

\IEEEPARstart{I}{n} many occasions, a textual description of our data (i.e., categorical data) is presented and no clear or precise labeling to perform a classification task. In order to deal with this type of data, a Data Mining technique called clustering is used. This techinique tries to segment data into different groups such as all samples inside a given group have higher similarity among themselves than with samples from other groups. As such, it is required that the data samples are comparable given a similarity metric.

For the categorical data, metrics that accounts for the number of common features shared by two samples are commonly used. One example is the Jaccard metric that, given the features sets of two objects, it calculates the ratio between the length of their intersection by the length of their union. One problem arises when we are dealing with data on a high-dimensional space since the probability of occurrence of common features between samples decreases, making it difficult to tell similar from dissimilar. This is called curse of dimensionality~\cite{har2012approximate} and it plagues many Machine Learning algorithms.

One way to deal with such problem is to transform the data into a lower dimensional representation while preserving enough information such as similar objects remains similar after the transformation. One of such methods is called Probabilistic Dimension Reduction~\cite{har2012approximate}. 

Probabilistic Dimension Reduction is performed exploiting the probability of two objects being very similar. This is done by means of specially designed hash functions that has a high probability of collision when hashing objects with similarity above a certain threshold. One of these techniques is called Minhashing, and it was specifically crafted to approximate the Jaccard Index between two sets. This is based on the fact that, given a random permutation of the features set, the probability that the first feature of two objects are the same is equal to the Jaccard Index between those two. In order to become computationally feasible, the random permutation is approximated by using an universal hash function.

But even so, using this dimension reduction technique may overlook \textit{hidden} relationships between two samples when they are similar under just a very small subset of features. Ideally, the clustering procedure must be performed in a two-way manner, i.e., by finding clusters of samples related to clusters of features. This clustering technique is called Co-Clustering.

Data Co-Clustering~\cite{franca2012scalable,dhillon2003information,labiod2011co,gaofast}, also known as biclustering, tries to find subsets of samples and features that maximizes the similarity of the select samples when considering the chosen features. It exploits the fact that a given sample may belong to different categories when viewed by different aspects of its description. For example, a given news text may report a story about the economies of a football team. This document will have terms that are related to sport and other terms that relates to economy. So, this sample will belong to two different classes when viewed by different features and, as such, must be compared differently for samples of different classes.

This technique allows several relaxations compared to the constraints of traditional clustering, such as, each object may belong to more than a single cluster, whenever they are viewed by different features. In the same way, a given feature may appear in different features clusters as it may have different meanings within different groups. Also, it lacks the need of specifying the number of clusters, since it simply tries to find every cluster with well-defined properties, i.e., clusters of samples sharing a subset of features.

Although this flexibility can be seem advantageous since it can extract many more knowledge from a data set, it can also result in some unintersting data relationship given the focus of the study. Also, the creation of algorithms with such flexibilities is not trivial, and such algorithms have implications that make it difficult to make a basis of comparison with the literature and give theoretical guarantees. As such, many Co-Clustering algorithms were created restraining some of these flexibilities by impeding the overlapping of samples and features among groups and pre-specifying the number of samples and features clusters~\cite{dhillon2003information,labiod2011co}.

Besides theses difficulties, there are some Co-Clustering algorithms capable of dealing with such flexibility, the most recents being the \textit{HBLCoClust}~\cite{franca2012scalable} and \textit{CoClusLSH}~\cite{gaofast}. They both have in common the use of a probabilistic dimension reduction technique, Locality Senstive Hashing, in order to pre-process the data set to find possible co-clusters.

Apart from that, they differ in how they utilize such probabilistic information and, thus, how they generate the final co-clusters set. \textit{HBLCoClust}, for example, depended on graph partitioning techniques in order to generate meaningful results, thus losing some of the aforemetioned flexibilities. In this paper a reformulation of this technique is proposed in order to keep its scalability while producing more informative co-clusters without depending on any kind of information, i.e., number of clusters. It will be shown in the next sections that this approach is capable of finding a high quality set of patterns while retained all of the flexibility expected from a Co-Clustering algorithm.

In Section~\ref{sec:cocluster} the Co-Clustering definition will be described in more details as well as some of its practical applications. Section~\ref{sec:hblcoclust} details the proposed algorithm detailing all of its key aspects and properties regarding the flexiblity of a co-clustering algorithm and scalability. Next, in Section~\ref{sec:experiments}, a complete set of experiments will be performed in order to assess how well the proposed algorithm performs on real-world scenarios against another similar Co-Clustering algorithm. A brief comparison with state-of-the-art algorithms will be given when approppriate in order to establish the advantages and disadvantages of Co-Clustering. Finally, Section~\ref{sec:conclusion} will comment on the concluding remarks together with some future perspectives.

\section{Co-Clustering}
\label{sec:cocluster}

Co-Clustering refers to the process of finding subsets of rows and columns from a data matrix such as the values of the extracted submatrix presents any kind of relationship~\cite{franca2012scalable,dhillon2003information,labiod2011co,gaofast,Hartigan1972,ChengChurch2000,Mirkin1996, de2010finding}. Usually, the rows and columns are described as objects, or samples, and features respectivelly from a Data Mining perspective. The relationship saught on this procedure will depend on the nature of the data set, and it can be submatrices with constant values, with constant values along the rows or along the columns, correlated values, values expressing any consistent ordering or dense submatrices from sparse data (see Fig.~\ref{fig:typebicluster}).

\begin{figure}
\begin{center}
\centering
	\subfigure[]{
		\footnotesize	
		$\left[
			\begin{array}{cccc}
				1 & 1 & 1 & 1\\
				1 & 1 & 1 & 1\\
				1 & 1 & 1 & 1
			\end{array} 
		\right]$	
	} \qquad
	\subfigure[]{
		\footnotesize
		$\left[
			\begin{array}{cccc}
				1 & 1 & 1 & 1\\
				2 & 2 & 2 & 2\\
				3 & 3 & 3 & 3
			\end{array}
		\right]$		
	}	\qquad	
	\subfigure[]{
		\footnotesize
		$\left[
			\begin{array}{cccc}
				1 & 2 & 3 & 4\\
				1 & 2 & 3 & 4\\
				1 & 2 & 3 & 4
			\end{array}  
		\right]$	
	}
		\subfigure[]{
		\footnotesize
		$\left[
			\begin{array}{cccc}
				3 & 2 & 1 & 5\\
				4 & 3 & 2 & 6\\
				2 & 1 & 0 & 4
			\end{array}  
		\right]$	
	}
\end{center}		
\caption[]{Example of different quality measures: (a) constant bicluster, (b) constant rows, (c) constant columns, (d) coherent values (rows or columns should exhibit high correlation).}
\label{fig:typebicluster}
\end{figure}

This procedure was already applied to a diverse set of applications such as gene expression analysis~\cite{de2010finding,de2008bicaco,MitraBanka2006,CoelhoEtAl2008b}, text mining~\cite{dhillon2003information,CastroEtAl2009}, recommendation systems~\cite{symeonidis2008providing,CastroEtAl2007b} and data imputation~\cite{de2013predicting}. The most popular variation of Co-Clustering algorithm are those applied to large gene expression data in order to find additive coherence~\cite{de2011extracting} in subsets of genes and conditions. Despite that, the application in the extraction of information of categorical data has been on the rise recently. In categorical data,  samples are described by the presence of some features from an usually large set of possibilities. This type of data can be described as a very sparse matrix where the present values indicate the presence of a feature on a sample.

An example of such data is text documents where each document can be described by the presence of a number of terms extracted from a large set of possible words. A Co-cluster of such data set would be a subset of documents that share a subset of words in common and, possibly, those words would describe the subject of these documents.

Formally, given the sets $O$ of objects and the set $F$ of features,  a Co-Cluster $C$ of a subset $O' \subset O$,  a subset $F' \subset F$ and a relation $R \subset O \times F$ can be described as:

\begin{equation}
  C(O',F') = \left\{O' \times F' \mid o \in O', f \in F', (o,f) \in R\right\},
\label{eq:hardcoclust}
\end{equation}

\noindent where $O'$ and $F'$ can also be denoted as the objects and features clusters respectively.

The relation $R$ contains all the tuples $(o,f)$ inside the dataset. Notice though that the constraint that every pair of object and feature on the co-cluster must exist in $R$ may be difficult to satisfy on very sparse data sets. For example, in text document data extracted from a news website, if we take three documents, one describing the victories of a football team, the other one telling about financial problems of this same team and the third about a fight among the fans of this team against fans of another one. Apart from common words, they will probably share just a short set of terms associated with the central theme. The more documents we try to insert into the document cluster the more likely one or more terms should be removed from the features cluster. 

In order to maximize the number of elements inside the clusters, without degrading the overall quality of a co-cluster,  the previous equation can be reformulated as:

\begin{equation}
C(O',F') = \left\{O' \times F' \mid |R'| \geq \rho \cdot |O' \times F'|\right\},
\label{eq:flexcoclust}
\end{equation}

\noindent where $|.|$ is the number of elements of a set and

\begin{equation}
R' = \left\{(o,f) \in R  \mid o \in O', f \in F'\right\}.
\end{equation}

Notice that there is no restriction whether an object or a feature must belong to a single cluster. There are other possible variations of this problem such as the $k,l-$Co-Clustering, in which it seeks to partition the data into $k$ objects clusters and $l$ features clusters by maximizing the number of non-null elements from the submatrice induce by each combination of $k$ and $l$. Specifically, in this work, the focus will be on the formulation given in Eq.~\ref{eq:flexcoclust}.

\section{Hash-based Linear Co-Clustering}
\label{sec:hblcoclust}

This section will first describe some fundamental theories and algorithms in order to better understand the proposed algorithm. First, the basics of probabilistic data mapping through random hash functions will be introduced. Following, an exact algorithm applied for a similar problem, called InClose, will be explained and, finally, the whole algorithm will be described in details.

\subsection{Locality Sensitive Hashing}  
\label{sec:minhashing}

A popular approach when dealing with high dimensional and high volume data sets is a probabilistic approach called Locality Sensitive Hashing (LSH). This algorithm exploits the fact that two very similar objects will likely collide when mapped through a weak hash function. In fact, depending on how this hash function is created, the probability of this collision is known to be proportional to their similarity.

One of such hash functions is the Minwise Independent Permutation (MinHash)~\cite{Carter1979143,har2012approximate} that states that the probability of collision of two objects is proportional to their Jaccard similarity. Jaccard Index or Jaccard Similarity is used when the data set is described through sets of categorical features. This similarity can be calculated as:

\begin{equation}
  J = \frac{\left|O1 \cap O2\right|}{\left|O1 \cup O2\right|},
\end{equation}

\noindent where $O1$ and $O2$ are the two objects compared and it returns a value between $0$ and $1$, with the former meaning the two objects are equal.

Minhash algorithm generates a random permutation $\pi$ of the features set and take the first feature of each object to be compared regarding the permutation order. The probability that these two elements are equal is given by:

\begin{equation}
P( O1_0^\pi = O2_0^\pi ) = \frac{\left|O1 \cap O2 \right|}{\left|O1 \cup O2\right|},
\end{equation}

\noindent that is equal to the Jaccard Index.

In order to have an estimate of such probability, it is possible to sample $M$ permutations and calculate the ratio in which the two features were the same. Notice that, for this to work, the permutation must follow an uniform distribution. Since generating a random permutation may be computationaly expensive, this can be approximated by using an universal hash function such as one of those proposed in~\cite{Carter1979143}:

\begin{equation}
h(x) = a \cdot x + b \mod P,
\label{eq:h}
\end{equation}

\noindent where $a$ and $b$ are randomly chosen with uniform distribution, and $P$ is a large prime number. The variable $x$ is the value to be hashed, i.e., a number associated with a given feature. This prime number should be at least as large as the number of features. This hash function will map each feature to an index in the range $[0,P[$. The random values will ensure that those indeces generate a random permutation of the feature set.

Notice that with this function the application of MinHash is straightforward. For each object $j$, simply find, for each hash function $i$, the feature $x$ that has the minimum value for $h_i(x)$:

\begin{equation}
mh_i(O_j) = \underset{x}{\arg\min} \left\{h_i(x) \mid \forall  x \in O_j\right\}.
\end{equation}

The complexity of this approach becomes $O(N.D.\bar{M})$ with usually $\bar{M} << M$ on sparse data sets. 

MinHash procedure made possible to go even further on optimizing the Nearest-Neighbor algorithm with the Locality Senstive Hashing~\cite{har2012approximate} procedure. This procedure creates $p$ hash keys for each object by grouping together a sequence of $k$ Minhashes and, thus, defining a bucket. Two objects, $o1$ and $o2$, will collied into the same bucket with probability:

\begin{equation}
P_{collision}(O1,O2) = 1 - (1 - J^{p})^{k})
\end{equation}

\noindent where $J$ is the desired Jaccard Index, $p$ is the number of hash functions per bucket and $k$ is the number of group of Minhashes.

So one can adjust the values of $k$ and $p$ in order to find all pair of documents with Jaccard above a given threshold with high probability.

\subsection{Co-Clustering enumeration with Formal Concept Analysis}
\label{sec:inclose}

In mathematics, there is a very similar field of study called Formal Concept Analysis~\cite{andrews2011close2}, which seeks the same relationship as described in Eq.~\ref{eq:hardcoclust} extracted from a boolean table. For such purpose they have devised algorithms that find the exact set of Formal Concepts, with exponential complexity, but with a reasonable time for data sets with thousands of objects and hundreds of features. 

This is done by exploiting the lexicographical order of the concepts so that not every possibility has to be evaluated. This algorithm, called \textit{InClose}~\cite{andrews2011close2} is summarized in Alg.~\ref{alg:inclose} and explained afterwards using the Co-Clustering notation for an easier reference.

\begin{algorithm}
  \SetKwInOut{Input}{input}\SetKwInOut{Output}{output}
\Input{dataset $D$, minimum size of object cluster $minO$ and minimum size of feature cluster $minF$}
\Output{set of Co-Clusters $C$}
\BlankLine
$Stack \leftarrow ( [0..m-1], \emptyset, 0 )$\;
\While{$Stack \neq \emptyset$}{
  $Oc, Fc, y \leftarrow pop(Stack)$\;
  $Candidates \leftarrow \emptyset$\;
  \For{$j \leftarrow y$ \KwTo $n-1$}{
    $Oc' \leftarrow Oc \cap \left\{i \mid D_{i,j} = 1\right\}$\;
    \If{$Oc' = Oc$}{
      $Fc \leftarrow Fc \cup \left\{j\right\}$\;
    }
    \ElseIf{length $Oc' \geq minO$ and $Oc'$ is canonnical }{
      $Fc' = Fc \cup {j}$\;
      $Stack \leftarrow Stack \cup \left\{ (Oc',Fc',j) \right\}$\;
    }
  }
  \If{length $Fc \geq minF$}{
    $C \leftarrow C \cup \left\{(Oc,Fc)\right\}$\;
  }
}
\caption{InClose}
\label{alg:inclose}
\end{algorithm}

In this algorithm, each candidate Co-Cluster is composed of a set of objects ($Oc$), a set of features ($Fc$) and an index indicating the last inserted feature ($y$). Initially, the algorithm creates a candidate with every object from the data set, an empty set of features and the first index ($0$), stacking it for further reference. This algorithm then goes through an iterative process until the stack is empty. 

Inside the main loop, it checks, for each feature from $y$ to the maximum index, what objects has a relation to it, forming a temporary object cluster $Oc'$. If this cluster is equal to the current object cluster ($Oc$), then this feature is inserted into the current feature cluster ($Fc$). Otherwise, if this set has a minimum number of objects and is considered cannonical (see below), then this set is inserted into the stack with this temporary object cluster, a feature cluster composed of the current one plus the current feature $j$ and with $j$ being the starting index.

Finally, after checking every feature, if the number of features on the feature cluster is above an established minimum, then the object cluster and feature cluster is inserted into a list of Co-Clusters ($C$).

An object cluster is considered cannonical if there is no other feature before $y$ that could be inserted into it without removing any object from the cluster.

\subsection{Proposed Algorithm}
\label{sec:hblcoclustalg}

The Hash-based Linear Co-Clustering algorithm, extended from~\cite{franca2012scalable}, connects the methods described on the previous subsections in order to generate an inexact set of Co-Clusters while maintaining computational scalability and filtering the most interesting patterns. The original algorithm was composed of three basic steps:

\begin{enumerate}
  \item Generate object clusters with the application of the LSH method, using the hash signature as the initial feature cluster.
  \item Expand these object clusters by inserting objects and features to its corresponding clusters through a simple local search.
  \item Induce a graph by connecting any two objects belonging to the same co-cluster and perform a community detection algorithm to find the final co-clusters set [optional].
\end{enumerate}

This third step, while optional, is performed in order to find fewer and larger co-clusters, while keeping them informative regarding the goal of the study. Without this step the algorithm produced too many small co-clusters impractical to post-analysis.

But, although the reported results were competitive, this last step made such algorithm dependable of a specific community detection algorithm. Also, it would restrain the algorithm to a $k,l$-co-clustering algorithm. In order to improve such algorithm and remove such constraints, a new general framework is proposed to it by following six steps:

\begin{enumerate}
  \item Pre-process the data set by removing any feature that appears more than a given threshold ($thr$) [optional].
  \item Perform the LSH method on the pre-processed data set to find the initial set of seed clusters.
  \item For each seed cluster, create a new data set composed of the union of the features set of each object and the union of the objects set of each feature.
  \item Perform the InClose algorithm to each of these datasets.
  \item For each generated co-cluster, insert objects and features that do not make it sparser than a threshold ($spthr$).
  \item Merge any two co-clusters that contains the same features cluster.
\end{enumerate}

The first step is optional and may be required for certain datasets that contains features that are related to most objects and may induce some uninteresting co-clusters into the data set. For example, in text documents there are many common words that have a very high probability of appearing, but do not describe the subject of such documents. In text mining this is usually dealt with by filtering through a stoplist, but this is not always possible when dealing with different sources of data, so a filtering threshold may be a better choice to generalize the algorithm.

In the second step, same as the original algorithm, it performs a probabilistic search on the dataset in order to find objects and features that are likely to be clustered, these clusters will be used as a seed for the next steps.

For each co-cluster seed, a new data set is generated formed by the subset of objects that relates at least to one of the seed features and the subset of features related to at least one of the seed objects. These new data sets will be generally small due to the sparse nature. So, the InClose algorithm is applied to them in order to find the exact set of co-clusters regarding these subset of data. Notice though that this step does not guarantee the exact solution for the original data.

Finally, in order to find significantly large co-clusters, the density contraint is relaxed and it inserts every object and feature that does not violate a maximum alowed sparsity threshold. As discussed before, this will lead to larger and less numerous co-clusters while maintaining their significance.

As a last step, since each co-cluster is generated from an incomplete view of the data set, the algorithm merge co-clusters that share the exact same feature set.

As the second steps depend on the features hashing signature, and since this algorithms is scalable regarding the data set size, as it will be shown in the next section, the five first steps are performed in the original data set and the transposed data before fine-tuning to the final solution.

Notice that beside the $thr$ and $spthr$ parameters, this algorithm also requires four others: the number of hashes to generate ($nhashes$), the number of hashes in each LSH ($nkeys$), and the minimum number of objects and features for each co-cluster ($minO$ and $minF$) for the InClose algorithm.

The overall algorithm is depicted in Alg.~\ref{alg:hblcoclust}.

\begin{algorithm}
  \SetKwInOut{Input}{input}\SetKwInOut{Output}{output}
  \Input{dataset $D$, minimum objects $minO$ and features $minF$, number of hashes $nhashes$, $nkeys$ groups of hashes, minimum feature count $thr$ and maximum sparsity $spthr$}
  \Output{set of Co-Clusters $C$}
\BlankLine
\tcc{Step 1: Pre-process data}
$D' \leftarrow \emptyset$\;
\For{$j \leftarrow 0$ \KwTo $n-1$}{
  \If{$\frac{\sum_{i}{D[i,j]}}{\sum{i,j}{D[i,j]}} < thr$}{
    $D' \leftarrow D' \cup D[:,j]$\;
  }
}
\tcc{Step 2: LSH}
$minhashes \leftarrow$ matrix of $nhashes$ for $o \in D'$\;
\For{$i \leftarrow 0$ \KwTo $m-n$}{
  \ForEach{$key$ in group of $nkeys$ from $minhashes$}{
        $clusters[key] \leftarrow clusters[key] \cup \left\{i\right\}$\;
  }
}
\tcc{Step 3, 4: Smaller datasets and InClose}
$C' \leftarrow \emptyset$\;
\tcc{The $value$ of an entry to $clusters$ will be a set of objects, while the $key$ is composed by a set of $nkeys$ features}
\ForEach{$(key, value)$ in $clusters$}{
  $Oc \leftarrow \left\{ o \in D' \mid \exists f \in key, (o,f) \in R\right\}$\;
  $Fc \leftarrow \left\{ f \in D' \mid \exists o \in value, (o,f) \in R\right\}$\;
  $C' \leftarrow C' \cup InClose(Oc,Fc)$\;
}
\tcc{Step 5: Insert elements}
$C \leftarrow \emptyset$\;
\ForEach{$(Oc,Fc)$ in $C'$}{
  \tcc{$R'$ is the set of relations in the co-cluster defined by $(Oc,Fc)$}
  $(Oc,Fc) \leftarrow \left\{(o,f) \in D' \mid |R'| > spthr\right\}$\;
  $C \leftarrow C \cup (Oc,Fc)$\;
}
\tcc{Repeat steps 2 to 5 with $D^{T}$}
\tcc{Step 6: Merge Co-Clusters}
\ForEach{ pair $(Oc1,Fc1), (Oc2,Fc2) \in C$ }{
  \If{$Fc1 = Fc2$}{
    \tcc{Merge both biclusters}
  }
}
\label{alg:hblcoclust}
\caption{HBLCOCLUST}
\end{algorithm}

The first three steps have a linear complexity regarding the data set size, the fourth step is exponential, but regarding the average size of the subset of rows. The last two steps have a quadratic complexity on the average co-cluster size.

\subsection{Comparative review}
\label{sec:literature}

In the literature the only work that can be directly compared with this approach, to the best of my knowledge, is the one called \textit{CoClusLSH}~\cite{gaofast}. This algorithm iteratively applies the LSH algorithm on objects and features, independently, while merging the found objects and features clusters by using an entropy metric. They tested this algorithm with a diverse set of real world data presenting competitive numerical results regarding purity, mutual information and number of groups found. They also found that their algorithm was scalable regarding data set size.

Another notable algorithm is the so called \textit{SpecCo}~\cite{labiod2011co} which reformulates the Co-Clustering problem as a graph partitionoing problem. It then applies a modularity maximization algorithm in order to find a $k,l$-co-clusters set. The presented results are also very positive but, unlike \textit{CoClusLSH} and \textit{HBLCoClust}, it does not scales linearly.

As a basis of comparison, the results obtained by \textit{HBLCoClust} algorithm will be compared with those obtained by \textit{CoClusLSH}, except in some justified cases. Additionaly, due to the unavailability of \textit{SpecCo} source code, the results presented here will be compared with those reported in~\cite{labiod2011co}, with proper observations.

\section{Experiments}
\label{sec:experiments}

In order to position the \textit{HBLCoClust} algorithm with other approaches in the literature, a set of experiments is devised in this section. The main goal of these experiments is to quantify the quality of the obtained co-clusters set as well as illustrate the practical applications of such clusters. With this purpose, data sets of different natures and applications were chosen from a set of widely used and publicly available.

The \textit{HBLCoClust} algorithm was implemented in Python 2.7 and the source code is available at~\url{https://github.com/folivetti/HBLCoClust} while \textit{CoClusLSH} was obtained from~\url{http://www.cs.sunysb.edu/~leman/pubs.html} and implemented in Matlab$\circledR$. Both algorithms was run under a Linux Debian 7.6 system on a i5-2450 @ $2.5$ GHz machine with $6$GB of RAM.

\subsection{Data sets and Applications}

The experiments are divided into $4$ different applications: categorical data clustering, text mining, topic modelling and recommender systems. 

Categorical data clustering is the process of finding clusters of data, described by categorical features, that shares common characteristics. When using a labeled data set, the majority of the objects on a given cluster is expected to be of the same category (purity). It should be noticed, though, that in data clustering, that is not possible to achieve in most of the time because all of the features have the same weight. As such, if most features point to a different classification then that of the given label, the purity of the clustering may be lower than the expected. In Co-Clustering application, clusters of elevated purity are expected, whenever the correct features are selected, and clusters of very low purity, whenever the selected features correspond to a different labeling.

For this application it was chosen the Zoo, House Votes $84$', Soybean small and Soybean large data sets~\cite{Bache+Lichman:2013}. The results obtained from these data sets were compared with \textit{CoClusLSH}, \textit{SpecCo} algorithms and the exact algorithm \textit{InClose}.

Text mining refers to the extraction of information from textual data, depending on the information available it may be used for topic classification, text clustering, topic modelling and natural language processing. In the same way as categorical data clustering, if a label is available, the clusters are expected to have a high value of purity. In the Co-Clustering scenario, some text documents will belong to different clusters, when considering a different set of features. Because of that, another possible quality measure is to calculate the pointwise mutual information of the features set to verify if they are more likely to co-occur than with any other feature outside this set.

The text data sets chosen for this experiment was the Classic-$3$~\cite{classic3} data set containing documents from $3$ different collections named CISI, CRAN and MED and two subsets of the $20$-newsgroups~\cite{Lang95} data sets, named here Multi5 containing texts from $5$ different newsgroups: comp.graphics, rec.sport.baseball, rec.motorcycle, sci.space, talk.politics.mideast, and Multi10 containing text from $10$ topics: alt.atheism, comp.sys.mac.hardware, misc.forsale, rec.autos, rec.sport.hockey, sci.electronics, sci.crypt, sci.med, sci.space, talk.politics.guns.

Topic modelling, one of the applications of text mining, refers to the process of finding the terms of a document that most describe its topics. It has many applications in summarization and finding meta-attibutes to text data. This application is explored by using the results from the Co-Clustering of Multi5 data set as it will be explained in the next section.

Finally, the Collaborative Filtering is a recommendation system approach that try to models the preference of each user for a serie of products through a rating matrix. This can be performed by different machine learning techniques such as nearest neighbor, clustering and matrix decomposition. The data for this application is usually descibed as a tuple $(user, item, rate)$ but this can be extended by including description of such items as well, so we would also have tuples of the type $(user, description, rate)$, where the rate could be the average rate of a given user for items containing this description. Notice that by doing this we would be dealing with two different data sets at the same time. This application was chosen to illustrate the flexibility of the proposed algorithm when dealing with data from different sources.

For this purpose it was used a combination of the Movielens data set~\cite{herlocker1999algorithmic} and the IMDB data avaliable at~\url{ftp://ftp.fu-berlin.de/pub/misc/movies/database/}.

Table~\ref{tab:datasets} summarizes the properties of each dataset used on these experiments.

\begin{table}[t]
\centering
\caption{Data set properties: number of objects, number of features, the number of relations (non-zero elements) and number of classes.}
\begin{tabular}{c|c|c|c|c}
\hline
& \textbf{\# obj.} & \textbf{\# feat.} & \textbf{Rel.} & \textbf{k} \\
\hline\hline
\textbf{Zoo} & $101$ & $16$ & $738$ & $7$ \\
\textbf{Soybean Small} & $47$ & $21$ & $880$ & $4$ \\
\textbf{Soybean Large} & $307$ & $35$ & $4 865$ & $19$ \\
\textbf{House Vote 84} & $435$ & $16$ & $6 568$ & $2$ \\
\textbf{Classic 3} & $3 891$ & $15 034$ & $227 355$ & $3$ \\
\textbf{Multi 5} & $5 000$ & $44 323$ & $539 933$ & $5$ \\
\textbf{Multi 10} & $10 000$ & $64 444$ & $987 443$ & $10$ \\
\textbf{Movielens} & $943$ & $4 233$ & $154 628$ & $-$ \\
\hline
\end{tabular}
\label{tab:datasets}
\end{table}

The adopted parameters for the \textit{HBLCoClust} are depicted in Table~\ref{tab:parameters} for each dataset. The \textit{InClose} parameters, $minO$ and $minF$ are the same adopted by \textit{HBLCoClust}. For the \textit{CoClusLSH} it was adopted the default parameters of $nhashes=100$ and $nkeys=3$ after empirically verified that these were optimal parameters regarding the adopted metrics and memory consumption.

\begin{table*}
\centering
\caption{Parameters used for \textit{HBLCoClust} on each data set.}
\begin{tabular}{c|c|c|c|c|c|c}
\hline
\textbf{dataset} & $\mathbf{minO}$ & $\mathbf{minF}$ & $\mathbf{nhashes}$ & $\mathbf{nkeys}$ & $\mathbf{thr}$ & $\mathbf{spthr}$ \\
\hline\hline
\textbf{Zoo} & $4$ & $6$ & $1 000$ & $2$ & $0.0$ & $1.0$ \\
\textbf{Soybean Small} & $4$ & $8$ & $1 000$ & $2$ & $0.1$ & $0.8$ \\
\textbf{Soybean Large} & $4$ & $10$ & $1 000$ & $2$ & $0.0$ & $0.8$ \\
\textbf{House Vote 84} & $10$ & $10$ & $1 000$ & $3$ & $0.4$ & $0.8$ \\
\textbf{Classic 3} & $50$ & $4$ & $2 000$ & $3$ & $0.2$ & $0.5$\\
\textbf{Multi 5} & $5$ & $5$ & $1 000$ & $3$ & $0.95$ & $0.5$ \\
\textbf{Multi 10} & $5$ & $5$ & $2 000$ & $3$ & $0.95$ & $0.5$ \\
\textbf{Movielens} & $2$ & $2$ & $5 000$ & $4$ & $0.0$ & $0.8$\\
\hline
\end{tabular}
\label{tab:parameters}
\end{table*}

\subsection{Metrics}

In order to quantify the quality of the obtained Co-Clusters, three metrics was chosen: Purity, Normalized Mutual Information and Pointwise Mutual Information.

Purity of a Co-Cluster measures the ratio between the number of the most frequent object label by the number of objects in the cluster. It essentialy quantifies for a given cluster if the majority of its objects are of the same type, i.e. the cluster is coherent regarding the labels:

\begin{equation}
  Purity(Oc) = \frac{ \max_{i}{|\left\{o \in Oc \mid l(o) = i\right\}|} }{ |Oc|  },
\end{equation}

\noindent where $l(.)$ is the label of a given object. This measure is averaged over all Co-Clusters.

Normalized Mutual Information calculates how likely is to find an object of a given label if a given co-cluster is selected at random. This metric is related to Purity but it also verifies the compactness of the Co-Clusters set, i.e. if the set has a minimum number of Co-Clusters. The highest possible value for this metric is impossible to achieve in some situations by the proposed algorithm, for example, when the data set contains two objects of the same label that do not share any feature. Ideally, for this metric, the Co-Clusters set should have the same number of clusters as the number of classes and every cluster has maximum purity:

\begin{equation}
  NMI = \frac{1}{H_{C}H_{l}} \sum_{c \in C, o \in Oc}{P( c, l(o) ) \times \log{\frac{ P( c, l(o) ) }{ P(c)P(l(o)) } } },
\end{equation}

\noindent where $H_C, H_l$ is the entropy of the Co-Clusters set and the labels, respectively.

Finally, the Pointwise Mutual Information measures the likelihood that the co-occurrence of any two features of a Co-Cluster was not by chance. This verifies if the subset of features selected by the Co-Cluster have significance regarding its corresponding objects:

\begin{equation}
  PMI = - \sum_{Of \in C} { \sum_{ f1,f2 \in Of} { \frac {\log{ \frac{P(f1,f2)}{P(f1)P(f2)}}}{\log{P(f1,f2)}}  }}.
\end{equation}

\subsection{Results}

The results for the first set of experiments, categorical data, are depicted in Table~\ref{tab:statsCat}~and~\ref{tab:resultsCat}. These Tables show the average results for $30$ runs of \textit{HBLCoClust} and \textit{CoClusLSH}, a single run of \textit{InClose} and the reported results for \textit{SpecCo}. The first table reports the number of found Co-Clusters, percentage of covered objects and features and average size of the Co-Clusters while the second reports the Purity, NMI and PMI.

From these tables the number of Co-Clusters found by \textit{HBLCoClust} is just a small fraction of possibilities enumerated by \textit{InClose}. Nonetheless, it is still capable of covering the entire set of objects. Since InClose only searches for dense Co-Clusters, some objects that contains less than $minF$ features may not be covered by the obtained set. Also, it is important to notice that \textit{CoClusLSH} gets closer to the number of classes than \textit{HBLCoClust} most of the time. Since the number of clusters is a parameter for \textit{SpecCo}, it will always return the desired number of classes. Regarding the number of features, \textit{HBLCoClust} not always cover the entire set, since it filters the uninteresting features it may not be capable of doing so.

Regarding the quality metrics, the first thing to notice is that \textit{CoClusLSH} obtained a worse result in every data set for Purity. This is due to its merging step that gives more importance for the reduction of the number of clusters than their sparsity. \textit{InClose} obtained the best purity for two of the four data sets, \textit{SpecCo} was the best in one of them and \textit{HBLCoClust} obtained one best result. Regardless, \textit{HBLCoClust} was always close to the best result.

When analysing the NMI, \textit{SpecCo} will always have the optimal number of clusters, regarding the data set classes, and thus its NMI will always be higher than the other approaches (except when it has a much lower purity). Disregarding the \textit{SpecCo} results, \textit{HBLCoClust} could obtain the best compromise between purity and number of clusters, having a higher value of NMI. Regarding PMI, \textit{HBLCoClust} obtained a higher value for two of those data sets, a close second place in one of them and significant worse result in one of the sets. Even so, it was capable of maintaining a positive value significantly above zero, thus correctly finding coherent set of features.

\begin{table}[t]
  \centering
  \caption{Statistics of obtained Co-Clusters set for the Categorical data sets.}
  \begin{tabular}{c|c|c|c|c}
    \hline
    \textbf{Zoo}        & \textbf{\#} & \textbf{Objs.} & \textbf{Feats.} & \textbf{Size} \\
    \hline\hline
    \textbf{HBLCoClust} & $27.07$     &  $1.00$        & $0.80$          & $83.07$       \\
    \textbf{CoClusLSH}  & $37.00$     &  $1.00$        & $1.00$          & $52.00$       \\
    \textbf{InClose}    & $67.00$     &  $0.81$        & $0.75$          & $114.00$       \\
    \textbf{SpecCo}     & $ 7.00$     &  $1.00$        & $--$            & $--$          \\
    \hline
    \textbf{Soybean S}        & \textbf{\#} & \textbf{Objs.} & \textbf{Feats.} & \textbf{Size} \\
    \hline\hline
    \textbf{HBLCoClust} & $13.80$     &  $1.00$        & $57.78$          & $83.13$       \\
    \textbf{CoClusLSH}  & $10.00$     &  $1.00$        & $1.00$          & $225.00$       \\
    \textbf{InClose}    & $225.00$     &  $1.00$        & $70.83$          & $121.00$       \\
    \textbf{SpecCo}     & $ 4.00$     &  $1.00$        & $--$            & $--$          \\
    \hline
    \textbf{Soybean L}        & \textbf{\#} & \textbf{Objs.} & \textbf{Feats.} & \textbf{Size} \\
    \hline\hline
    \textbf{HBLCoClust} & $42.40$     &  $1.00$        & $0.74$          & $205.73$       \\
    \textbf{CoClusLSH}  & $20.00$     &  $1.00$        & $1.00$          & $1 089.00$       \\
    \textbf{InClose}    & $6 470.00$     &  $0.98$        & $0.93$          & $109.00$       \\
    \textbf{SpecCo}     & $19.00$     &  $1.00$        & $--$            & $--$          \\
    \hline
    \textbf{House Votes}        & \textbf{\#} & \textbf{Objs.} & \textbf{Feats.} & \textbf{Size} \\
    \hline\hline
    \textbf{HBLCoClust} & $23.30$     &  $1.00$        & $0.85$          & $1 173.80$       \\
    \textbf{CoClusLSH}  & $18.00$     &  $1.00$        & $1.00$          & $734.00$       \\
    \textbf{InClose}    & $12 4371$     &  $0.95$        & $1.00$          & $225.00$       \\
    \textbf{SpecCo}     & $ 2.00$     &  $1.00$        & $--$            & $--$          \\
    \hline
  \end{tabular}
  \label{tab:statsCat}
\end{table}

\begin{table}[t]
  \centering
  \caption{Obtained results for the categorical data sets.}
  \begin{tabular}{c|c|c|c}
    \hline
    \textbf{zoo}        & \textbf{Purity} & \textbf{NMI} & \textbf{PMI} \\
    \hline\hline
    \textbf{HBLCoClust} & $0.88$     &  $0.29$        & $0.18$ \\
    \textbf{CoClusLSH}  & $0.79$     &  $0.25$        & $0.29$ \\
    \textbf{InClose}    & $0.93$     &  $0.19$        & $0.21$ \\
    \textbf{SpecCo}     & $0.90$     &  $0.92$        & $--$  \\
    \hline
    \textbf{Soybean S}        & \textbf{Purity} & \textbf{NMI} & \textbf{PMI} \\
    \hline\hline
    \textbf{HBLCoClust} & $0.89$     &  $0.40$        & $0.25$ \\
    \textbf{CoClusLSH}  & $0.56$     &  $0.30$        & $0.08$ \\
    \textbf{InClose}    & $0.59$     &  $0.09$        & $-0.26$ \\
    \textbf{SpecCo}     & $1.00$     &  $1.00$        & $--$  \\
    \hline
    \textbf{Soybean L}        & \textbf{Purity} & \textbf{NMI} & \textbf{PMI} \\
    \hline\hline
    \textbf{HBLCoClust} & $0.73$     &  $0.26$        & $0.15$ \\
    \textbf{CoClusLSH}  & $0.28$     &  $0.09$        & $0.06$ \\
    \textbf{InClose}    & $0.50$     &  $0.07$        & $0.13$ \\
    \textbf{SpecCo}     & $0.67$     &  $0.78$        & $--$  \\
    \hline
    \textbf{House Votes}        & \textbf{Purity} & \textbf{NMI} & \textbf{PMI} \\
    \hline\hline
    \textbf{HBLCoClust} & $0.91$     &  $0.29$        & $0.43$ \\
    \textbf{CoClusLSH}  & $0.85$     &  $0.24$        & $0.45$ \\
    \textbf{InClose}    & $0.93$     &  $0.10$        & $0.26$ \\
    \textbf{SpecCo}     & $0.87$     &  $0.47$        & $--$  \\
    \hline
  \end{tabular}
  \label{tab:resultsCat}
\end{table}

For the second set of experiments, textual data, it was not possible to run neither \textit{InClose}, due to computational complexity limits nor \textit{CoClusLSH} due to a memory constraint. In this situation we are dealing with much larger data sets that most algorithms require some adaptation to deal with. To give us a comparison baseline these results will be compared to those reported by~\cite{labiod2011co} for the \textit{SpecCo} algorithms on similar datasets: C150, composed of $150$ documents of Classic3 and $3 625$ words, NG5 and NG10, both composed of $500$ documents extracted from Multi5 and Multi10, respectively, and containig $2 000$ words. These words features were selected by the authors with a supervised approach.

In this situation, \textit{HBLCoClust} will deal with a much larger object set and will blindly select the least frequent words of each data set according to $thr$ parameter.

From Tables~\ref{tab:statsText}~and~\ref{tab:resultsText},  \textit{HBLCoClust} still maintains a much larger set of Co-Clusters but, the Purity obtained by \textit{HBLCoClust} was the same as the reported value for \textit{SpecCo} for the Classic3 and much superior when regarding the Multi5 and Multi10 data sets. When comparing NMI, again the \textit{SpecCo} has the advantage of having the correct number of clusters, so it obtained a much higher value. The features sets obtained by \textit{HBLCoClust} presented a considerable high value of PMI, meaning that they are coherent regarding their corresponding documents sets.

\begin{table}[t]
  \centering
  \caption{Statistics of obtained Co-Clusters set for the Textual data sets.}
  \begin{tabular}{c|c|c|c|c}
    \hline
    \textbf{Classic3}        & \textbf{\#} & \textbf{Objs.} & \textbf{Feats.} & \textbf{Size} \\
    \hline\hline
    \textbf{HBLCoClust} & $219.20$     &  $1.00$        & $0.60$          & $2 807.23$       \\
    \textbf{SpecCo}     & $ 3.00$     &  $1.00$        & $--$            & $--$          \\
    \hline
    \textbf{Multi5}        & \textbf{\#} & \textbf{Objs.} & \textbf{Feats.} & \textbf{Size} \\
    \hline\hline
    \textbf{HBLCoClust} & $1 054.97$     &  $1.00$        & $0.18$          & $1 051.23$       \\
    \textbf{SpecCo}     & $ 5.00$     &  $1.00$        & $--$            & $--$          \\
    \hline
    \textbf{Multi10}        & \textbf{\#} & \textbf{Objs.} & \textbf{Feats.} & \textbf{Size} \\
    \hline\hline
    \textbf{HBLCoClust} & $2 819.33$     &  $1.00$        & $0.16$          & $661.00$       \\
    \textbf{SpecCo}     & $10.00$     &  $1.00$        & $--$            & $--$          \\
    \hline
  \end{tabular}
  \label{tab:statsText}
\end{table}

\begin{table}[t]
  \centering
  \caption{Obtained results for the Textual data sets.}
  \begin{tabular}{c|c|c|c}
    \hline
    \textbf{Classic3}        & \textbf{Purity} & \textbf{NMI} & \textbf{PMI} \\
    \hline\hline
    \textbf{HBLCoClust} & $0.86$     &  $0.14$        & $0.20$ \\
    \textbf{SpecCo}     & $0.86$     &  $0.73$        & $--$  \\
    \hline
    \textbf{Multi5}        & \textbf{Purity} & \textbf{NMI} & \textbf{PMI} \\
    \hline\hline
    \textbf{HBLCoClust} & $0.91$     &  $0.18$        & $0.37$ \\
    \textbf{SpecCo}     & $0.59$     &  $0.53$        & $--$  \\
    \hline
    \textbf{Multi10}        & \textbf{Purity} & \textbf{NMI} & \textbf{PMI} \\
    \hline\hline
    \textbf{HBLCoClust} & $0.82$     &  $0.14$        & $0.33$ \\
    \textbf{SpecCo}     & $0.57$     &  $0.55$        & $--$  \\
    \hline
  \end{tabular}
  \label{tab:resultsText}
\end{table}

These obtained values of PMI for the textual data is very close to the PMI obtained by topic modelling approaches regarding a set of words describing the topic of each document~\cite{anandkumar2013learning}, with approximately the same size of features clusters (hidden variables in topic modelling). This creates the possibility of using the features set as a set of topics describing a given document. To illustrate this assumption, Table~\ref{tab:topics} contains the terms of $5$ features clusters chosen at random for the Multi5 data set each one pertaining to one of the data set classes.

From this Table, almost every term associated with a cluster clearly identifies the corresponding classification of the documents. This application will be explored in further details on future research.

\begin{table*}
  \centering
  \caption{Terms extracted from features clusters of Multi5 data set.}
  \begin{tabular}{c|p{13cm}}
    \hline
    \textbf{sport.baseball} & reds, houston, standings, cincinnati, colorado, mets, scores, marlins, including, milwaukee, oakland, pirates, city, expos, los, west, indians, minnesota, ocf, rangers, joseph, white, angels, texas, giants, toronto, pittsburgh, phillies, cardinals, cubs, atlanta, mariners, orioles, mlb, lost, braves, louis, detroit, teams, athletics, streak, hernandez, san, boston, cleveland, dodgers, sox, seattle, astros, blue, diego, jays, jtchern, rockies, twins, brewers, tigers, red, francisco, philadelphia, kansas, yesterday, royals, california, padres, berkeley, league, chicago, florida, angeles, april, montreal, yankees, baltimore, york \\
    \textbf{motorcycles} & handlebars, motorcycle, speed, countersteering, foward, handle, faq, awful, turns, debating, ummm, uiuc, happens, turning, pushing, fgc, convert, unb, explain, duke, unbvm, acpub, slack, zkcl, infante, cbr, eric, cso, csd, methinks, push \\
    \textbf{politics.mideast} & later, muslims, ohanus, vol, turkish, involved, roads, argic, hand, muslim, armenian, document, russian, armenians, including, army, sahak, proceeded, serdar, soul, killed, among, children, published, blood, appressian, mountain, often, exists, turks, armenia, general, soviet, serve, escape, genocide, melkonian, ways, extermination, passes, closed  \\
    \textbf{sci.space} & six, rigel, aurora, wings, mary, dfrf, alaska, dryden, spin, military, facility, speak, unknown, digex, prb, flight, pilot, fly, kotfr, air, nsmca, pat, edwards, mig, wire, fighter, shafer  \\
    \textbf{comp.graphics} & plot, recommend, pascal, hidden, routines, object, basic, cost, address, bob, cad, mac, info, frame, short, offer, animation, price, sites, across, package, low, building, directory, removal, documentation, robert, built, recommendations, libraries, various, tasks, shading, fast, files, code, objects, tools, handle, demo, library, contact, book \\
    \hline
  \end{tabular}
  \label{tab:topics}
\end{table*}

Finally, as a last experiment, a simple recommendation system was devised by using the information provided by a Co-Clusters set. In order to create a categorical data set from the rating data provided by the Movielens Data Set, the tuple (user, movie, rating) was changed into (user, movieY) if the rating given by the user to the movie was higher than $2$ and (user, movieN) otherwise.
Additionally, some meta-attributes were extracted for each movie on this data set by using the IMDB interface data. The chosen attributes were: genre, actors, actress, directors and keywords. Each attribute was then related to each user by averaging the ratings given by a user to a movie containing this attribute and the averaged rating was converted as in the movies tuples, so tuples with the format (user, attributeY) and (user, attributeN) were created.

This generated a data set containing $943$ users, $4 233$ attributes and $154 628$ relations extracted from $80 000$ ratings from the Movielens training set. \textit{HBLCoClust} algorithm was applied to the positive (Y) and negative (N) relations separatly in order to create a \textit{like} and \textit{dislike} ruleset for each users cluster.

Afterwards, for each user it was created a \textit{like} profile containing every feature of every positive co-cluster this user belonged to and, similarly, a \textit{dislike} profile from the negative co-clusters. The intersection of these two sets were then removed from each profile.

For each of the remaining $20 000$ tuples (user, movie, rating) from the Movielens test set, the attributes of each movie was extracted in the same way as described above and the Jaccard Index of the meta-attributes with the \textit{like} and \textit{dislike} profile of this user was calculated, the predicted rating is the one profile most similar.

Since not every movie was covered by the Co-Clusters, only $12 862$ test ratings could be predicted. From this total, this simple approach obtained an accuracy of $83.14\%$, a well-known approach from the literature, regularized SVD~\cite{funk2006netflix}, obtained $79.68\%$ from this same set of predicted ratings. Another approach, Naive Bayes~\cite{lewis1998naive}, achieved $69.32\%$ of accuracy.

Using the entire test set for the sake of comparison, Naive Bayes achieved $79.2\%$ of accuracy while regularized SVD achieved $66.01\%$. Besides obtaining a higher accuracy, the Co-Clustering technique also pre-select only the movies it has sufficient information to form a co-cluster. While this might limits the amount of possible predictions it also might tip on what predictions are more relevant. Also, the information given by the profiles can enrich the recommendation experience by providing an explanation for them~\cite{symeonidis2008providing}. This experiment still requires a more detailed investigation and as such will be the subject of a future research.

In Table~\ref{tab:taste} the profiles of a given user, as well as some of the movies rated by him is depicted.

\begin{table*}
  \centering
  \caption{Profile generated for one of the user on Movielens data set.}
  \begin{tabular}{c|p{13cm}}
    \textbf{Movies} & Jaws, Back to the Future, Twelve Monkeys, Dumb \& Dumber, ... \\
    \textbf{like} & disaster, infidelity, horse, gunfight, USA, automobile, hospital, bathroom, jealousy, racism, elevator, fight, beer, male-nudity, helicopter, impalement, good-versus-evil, outer-space, murder, washington-d.c., fire, shot-to-death, los-angeles-california, independent-film, small-town, train, drunkenness, one-man-army, baby, teenage-boy, lifting-someone-into-the-air, redemption, f-word, photograph, tough-guy, gangster, main-character-dies \\
    \textbf{dislike} & second-part, beaten-to-death, haunted-by-the-past, Washington,DistrictofColumbia,USA, cell-phone, vengeance, bulletproof-vest, obsession, book, die-hard-scenario\\
  \end{tabular}
  \label{tab:taste}
\end{table*}

\begin{figure}
  \centering
  \includegraphics[trim = 5mm 0mm 0mm 0mm, clip, width=0.49\textwidth]{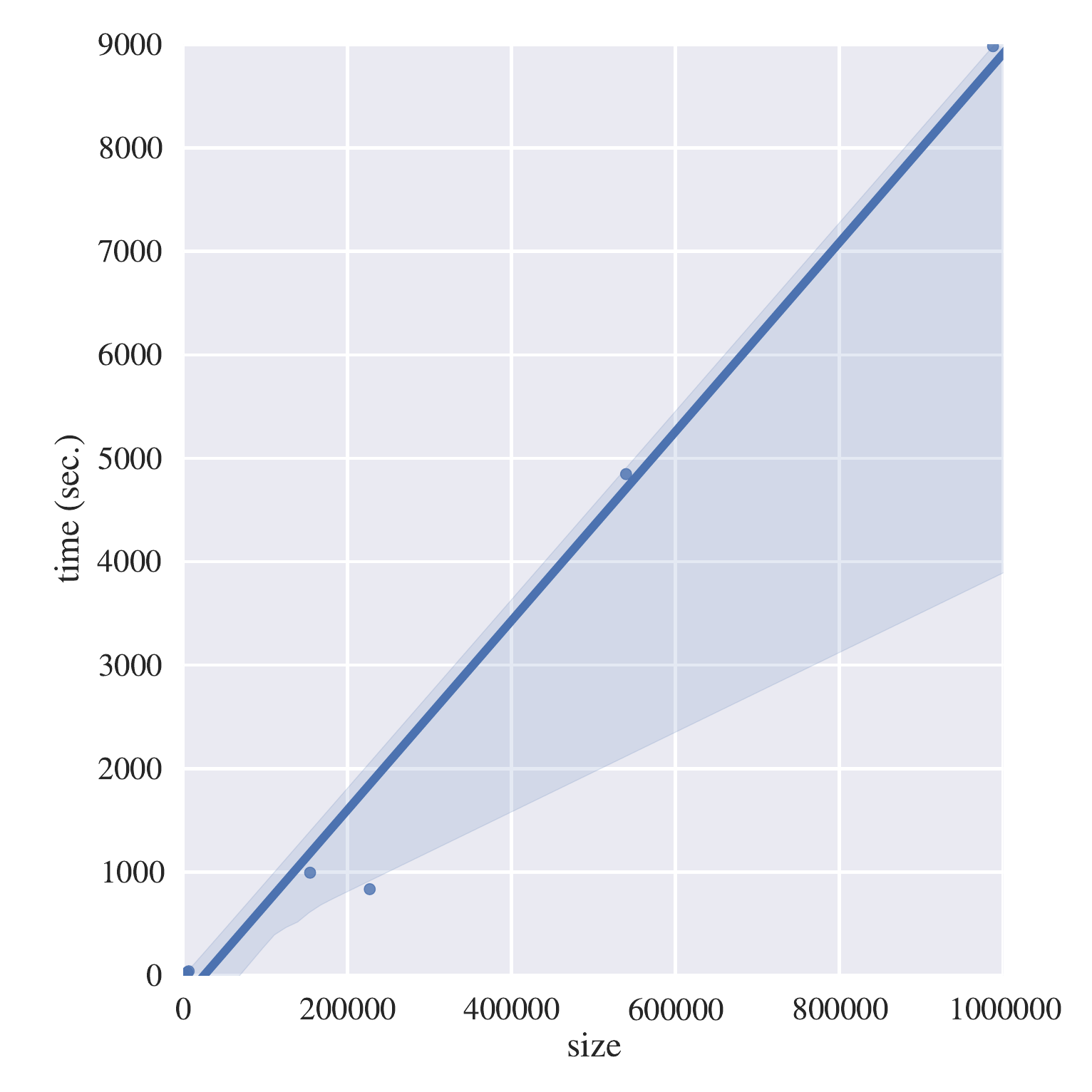}
  \caption{Time complexity estimation.}
  \label{fig:time}
\end{figure}

Regarding the time complexity of this algorithm, each step is proportional to the data size outputed by the previous step. So, in order to verify its complexity regarding the number of relations contained in the data set, the algorithm was run with for each of the presented data sets with a fixed parameters set: $nhashes=1000$, $nkeys=3$, $minO=minF=4$, $thr=0.0$, $spthr=1.0$. The result is shown in Fig.~\ref{fig:time} together with a regression line. It can be seen from this figure that this algorithm seems to have indeed a linear complexity regarding data set size.

\section{Conclusion}
\label{sec:conclusion}

In this paper a new algorithm for categorical data Co-Clustering was presented extending the author's previous approach. This algorithm consists of six sequential steps and scales linearly with the number of non-empty entries of the data set. This is achieved by using a probabilistic approach to approximate the partial similarity between objects, thus creating \textit{seed} co-clusters. These seed co-clusters are used to select a subset of the whole data set allowing the use of an enumerative algorithm called \textit{InClose}, afterwards it merges those co-clusters found by the different views of the data set.

The experiments performed on categorical data clustering and text mining showed that this approach stays \textit{on par} with others Co-Clustering approaches on small data sets and outperforms them when these data sets become larger. Also, unlike the other approaches, the corresponding features cluster of each co-cluster maintened a high value of pointwise mutual information, leading to meaningful features set that explains each cluster.

Examples of the usefulness of these features clusters were provided by means of two applications: topic modelling and collaborative filtering. In the first, it was shown that the selected features of a given cluster correctly and coherently describes the grouped topic. In the second example, not only a higher accuracy was obtained when compared to traditional approaches but also each user was given a profile of its taste, thus explainning why a movie is recommended.

For the future researches following this paper, the previous applications examples will be further explored with a more thorough set of experiments and positioning among the state-of-the-art in each of these problems. Also, a further inspection of the usefulness of the features cluster will be investigated together with other applications that may be suitable to a Co-Clustering algorithm.



%% References


\bibliographystyle{IEEEtran}
\bibliography{HashBasedCoClustering}






% \end{thebibliography}

\begin{IEEEbiography}{Fabr\'{i}cio Olivetti de Fran\c{c}a}
received his Dr.E.E. degree from the University of Campinas (Unicamp), Campinas, SP, Brazil, in 2010. His main research areas are computational intelligence, ant systems, clustering, co-clustering, dynamic, combinatorial and multiobjective optimization, artificial immune systems, collaborative filtering and recommendation systems.
\end{IEEEbiography}
\end{document}